\documentclass{styles/svproc}
\usepackage{url}
\usepackage{amsmath,amssymb}
\usepackage{url}
\usepackage{cite}
\usepackage{caption}
\usepackage{subcaption}
\usepackage{graphicx} 

\usepackage{comment}
\usepackage[ruled,vlined]{algorithm2e}
\usepackage{enumitem}

\begin{document}
\mainmatter              
\title{Extended Hybrid Zero Dynamics for Bipedal Walking of the Knee-less Robot SLIDER}


%
\titlerunning{Extended HZD for Bipedal Walking of SLIDER}  
%
\author{Rui Zong\inst{1} \and Martin Liang\inst{2}, Yuntian Fang\inst{3},
Ke Wang\inst{1} \and Xiaoshuai Chen\inst{1} \and Wei Chen\inst{1} \and Petar Kormushev\inst{1}}
\authorrunning{Rui Zong et al.} 
%
\tocauthor{}
\institute{Robot Intelligence Lab, Dyson School of Design Engineering,\\
Imperial College London, UK\\\email{rui.zong21@imperial.ac.uk},\\
\and
Department of Electrical and Electronic Engineering,\\
Imperial College London, UK\\
\and
Department of Aeronautics, Imperial College London, UK\\
}

\maketitle              

\begin{abstract}

Knee-less bipedal robots like SLIDER have the advantage of ultra-lightweight legs and improved walking energy efficiency compared to traditional humanoid robots.
In this paper, we firstly introduce an improved hardware design of the SLIDER bipedal robot with new line-feet and more optimized mass distribution that enables higher locomotion speeds.
Secondly, we propose an extended Hybrid Zero Dynamics (eHZD) method, which can be applied to prismatic joint robots like SLIDER.
The eHZD method is then used to generate a library of gaits with varying reference velocities in an offline way. 
Thirdly, a Guided Deep Reinforcement Learning (DRL) algorithm is proposed to use the pre-generated library to create walking control policies in real-time. 
This approach allows us to combine the advantages of both HZD (for generating stable gaits with a full-dynamics model) and DRL (for real-time adaptive gait generation). The experimental results show that this approach achieves 150\% higher walking velocity than the previous MPC-based approach.

Video: \url{https://www.imperial.ac.uk/robot-intelligence/robots/slider/}

\end{abstract}

\noindent\textbf{Keywords:} Bipedal Robot, Hybrid Zero Dynamics, Reinforcement
Learning.

\section{Introduction}
Bipedal robots have promising applications in human-centered environments including assistive care and agile locomotion. The vast majority of bipedal robots, e.g., \cite{Hardware}, are built with full revolute joints—a design approach that introduces singularity issues when the knees are fully extended during walking. Additionally, knee actuators contribute significant weight and increase the inertia of the legs.
To address this challenge, knee-less prismatic-jointed robots such as SLIDER\cite{Wang_IROS-2020}\cite{Wang_Design} have been optimized with lightweight leg designs, which in turn significantly reduce the mass and inertia during walking. Due to its inherent linear structure that functions as an impact cushion, \cite{SLIDER2} employed Bayesian optimization to demonstrate the energy savings of the prismatic joint across a range of frequencies.

However, similar studies have lacked system stability considerations over the entire dynamic system, i.e., the planar contact constraints of the entire supporting foot are needed to allow the robot to track reference trajectories during its swing phase\cite{MPC0}. A static, precisely matched reference trajectory does not imply that the trajectory can still be optimal in the dynamic system. In fact, the study of the long-horizon of the state of a robotic system shows that even if a reference trajectory is initially optimal under static conditions, its performance degrades over time due to the accumulation of dynamic effects, uncertainties, and disturbances\cite{Long_Horizon}.
It explain that why the long-horizon hybrid methods of learning and model-based control in manipulation\cite{Manipulation} and the long-horizon walking trajectory optimisation such as Model Predictive Control (MPC)\cite{MPC0}\cite{MPC1} have become increasingly popular.
For robot locomotion, it is insufficient to design a controller solely on aligning the current state with the system's linearized asymptotic stability point, especially when many robotic systems do not exhibit such a stability point at all\cite{HZD0}.

\begin{figure}[h]
    \centering
    \begin{subfigure}{0.23\textwidth}
        \centering
        \includegraphics[width=\textwidth]{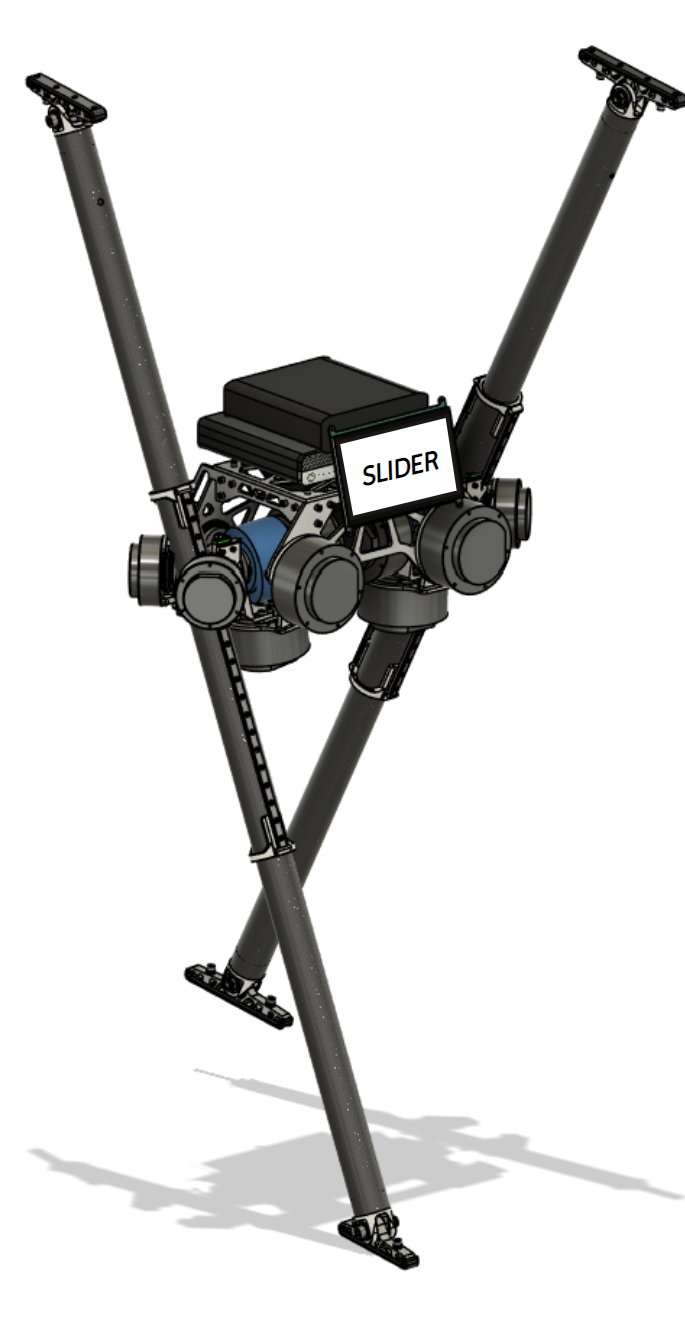}
        \label{fig:image1}
        \caption{}
    \end{subfigure}
    \begin{subfigure}{0.17\textwidth}
        \centering
        \includegraphics[width=\textwidth]{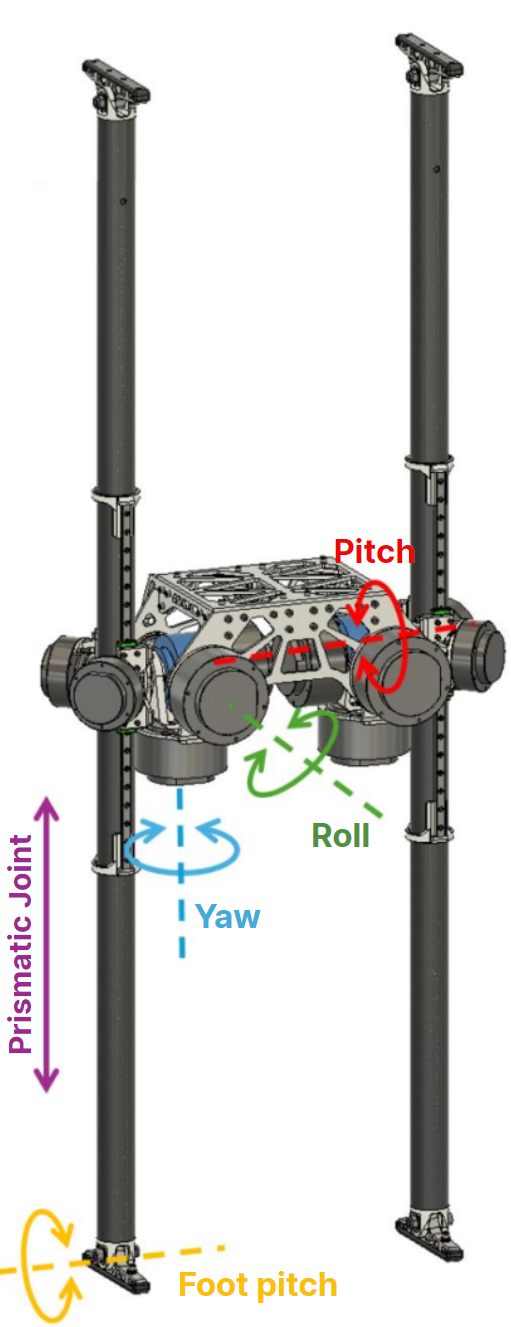}
        \label{fig:twinb}
        \caption{}
    \end{subfigure}
    \begin{subfigure}{0.22\textwidth}
        \centering
        \includegraphics[width=\textwidth]{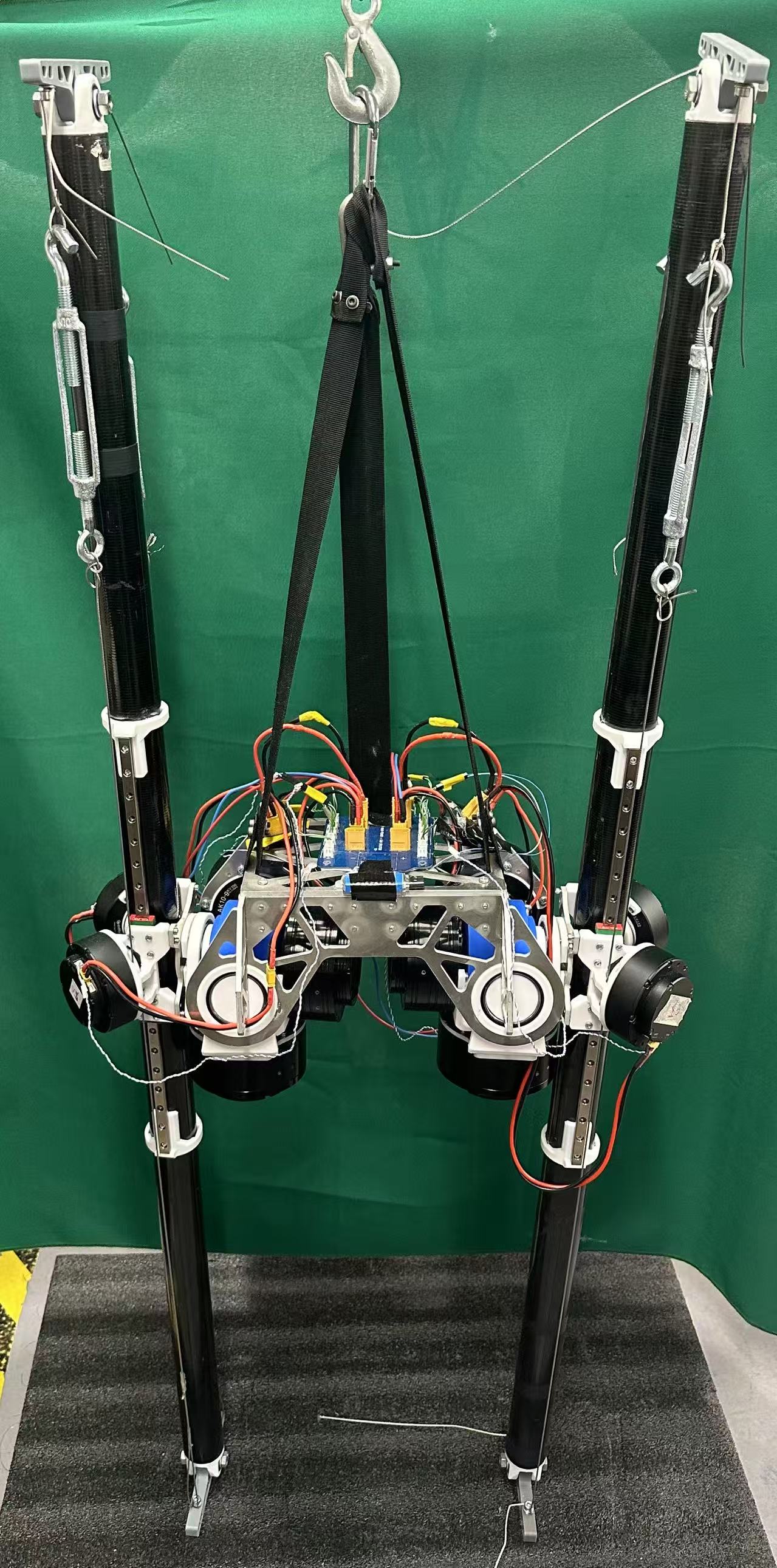}
        \label{fig:image3}
        \caption{}
    \end{subfigure}
    \caption{Improved SLIDER robot, (a) is the walking model of SLIDER, (b) describe the joints of the SLIDER, while (c) shows the SLIDER in the real world}
    \label{fig:combined}
\end{figure}

We propose an improved SLIDER structure, shown in Fig.~\ref{fig:combined}, which is a continuation of the previous knee-less design of SLIDER\cite{Wang_IROS-2020}, by incorporating a line-foot and optimized mass distribution. The line-foot greatly reduces the weight of the robot's foot, resulting in a significant reduction in the amount of energy consumed by the foot in the traditional design. At the same time, because of its line contact nature, the motion will make full use of the angular momentum generated by the lateral gravity, which increases the overall flexibility of the motion. We mounted a new Yaw joint on the hip, which enhances the redundancy of the whole system. All actuators are mounted in the the center of the hip, which further significantly reduce the leg weight.

Different control schemes have been used with SLIDER \cite{MPC0} and similarly structured bipedal robots. However, due to the nature of the underactuated system\cite{HZD1}, it is difficult to provide flexible locomotion control by purely optimal control. Therefore, we propose eHZD for offline trajectory generation. 
HZD\cite{HZD0} is an approach in which, instead of analyzing asymptotic equilibrium, a Poincare map is analyzed to obtain a limit cycle. It
 restrict robots' dynamics to a lower-dimensional space through virtual constraints and directly constructs all the motion processes of the whole nonlinear system into a large nonlinear optimisation problem. Although it takes into account the motion of both continuous and discrete systems, it can only be applied to robots with fully revolute joints\cite{HZD1}. In this paper, we propose an extended HZD method which can generate an approximate revolute model to capture a gait library at different speeds in a prismatic model. Even though the gait library considers all dynamic information through the whole locomotion part, it requires a low-noise environment and the motion pattern is rigid. Thus, a Guided DRL framework similar to \cite{DRL1} for the eHZD is proposed for adaptive real-time control policy generation.

The novelty of this paper lies in the following:\\[-0.8em]

\begin{enumerate}[topsep=0pt, partopsep=0pt, itemsep=0pt, parsep=0pt]
\item Improved SLIDER model with line-foot, yaw joint, and highly centralized mass.
\item HZD is extended to be applicable to robot models with prismatic joints. 
\item A modified Guided DRL control policy based on the obtained gait library is proposed.
\end{enumerate}

\section{Improved SLIDER Model Design}

\subsection{Mass Reduction in Base Link and Child Joints}
\begin{figure}
    \centering
    \begin{subfigure}{0.55\textwidth}
        \includegraphics[width=\linewidth]{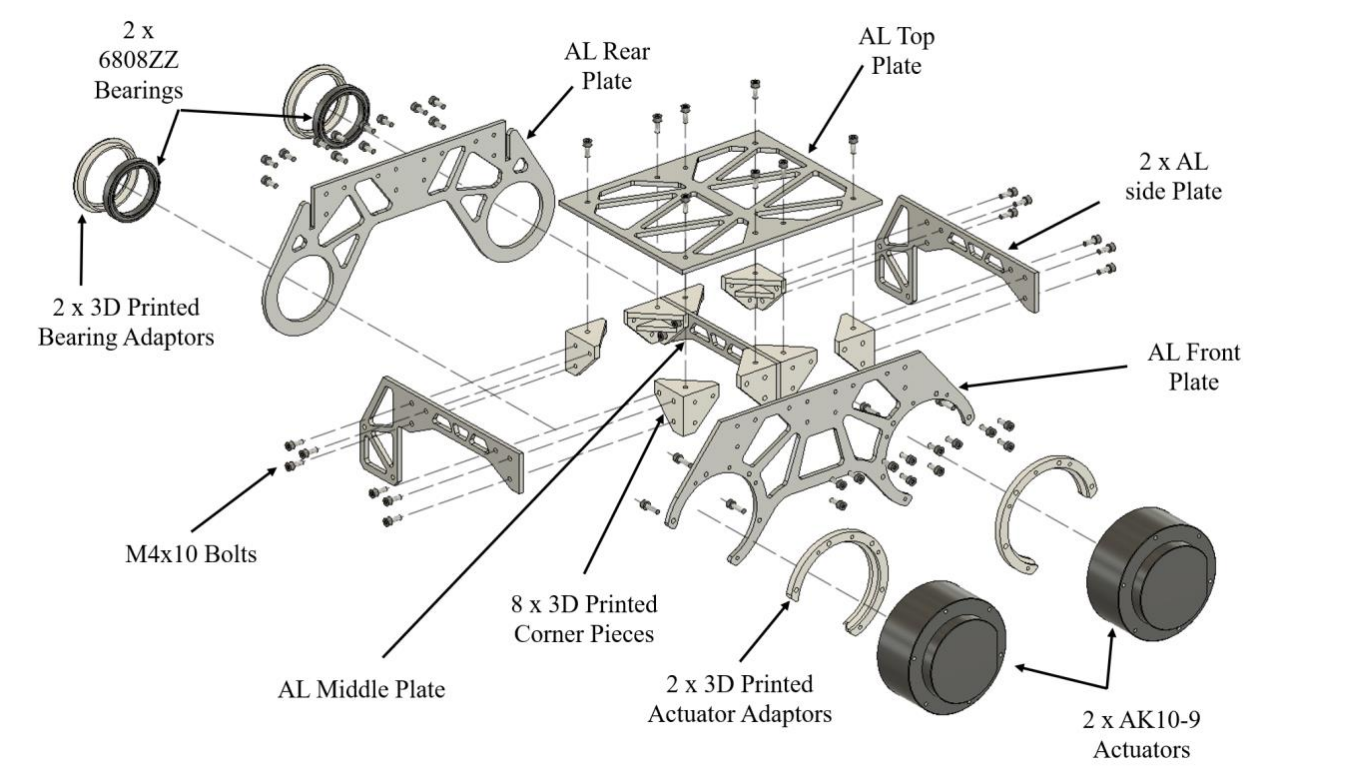}
        \caption{}
        \label{fig:exp_1}
    \end{subfigure}
    \hfill
    \begin{subfigure}{0.44\textwidth}
        \includegraphics[width=\linewidth]{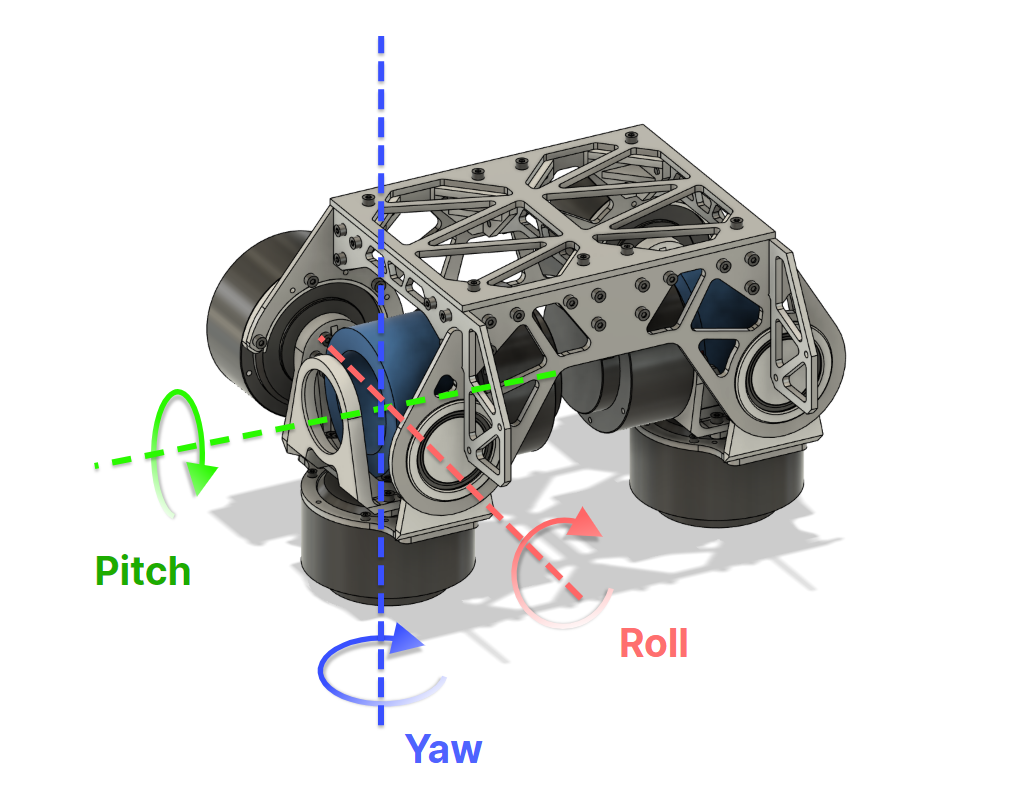}
        \caption{}
        \label{fig:asem_b}
    \end{subfigure}
    \caption{Improved Base link design: (a) Exploded view of the frame structure, (b) Base link assembly with child joints }
    \label{fig:main}
\end{figure}

In the previous generation, the SLIDER’s main body (Baselink in the URDF joint structure) used BLDC motors with integrated planetary gearboxes and FOC drivers, offering high torque density and precision feedback. However, the assembly was bulky and heavy due to standard aluminum extrusions and high-infill 3D-printed mounts. To optimize for extreme conditions, the new design uses high-performance materials and custom motor mounts. The main body is now constructed from 6082T6 aluminum alloy plates, chosen for their high stiffness-to-weight ratio, ease of manufacturing, and low cost. A panelized joint structure assembles the 3D frame from 2D sheet cut-outs, forming a truss. Mortise-and-tenon joints and 3D-printed reinforcements secure connections, while a 4mm aluminum sheet ensures rigidity, support self-tapping screws and enhances expandability. The overall assembly and exploded view are shown in Fig~\ref{fig:exp_1}.
The subsequent child joints, including the yaw and roll joints, were redesigned with optimized, rigid 3D-printed parts featuring reinforcement ribs for enhanced strength in a compact form factor. Additionally, to minimize the radial load on the motor output shaft, both the roll and yaw axes adopt a coaxial mounting design. By integrating an additional bearing at the load's distal end, the moment load on the motor's internal output shaft bearing is reduced, improving motor lifespan and structural robustness. The overall assembly is illustrated in Fig.~\ref{fig:asem_b}, with the name of each joint axis labeled.
\subsection{Differential Capstan Drive for Foot and Prismatic Sliding Joints Actuation}
  
The SLIDER’s prismatic joint, with its long travel range and centralized stationary actuators, requires precise linear constraint and effective torque transmission, which is particularly challenging due to the carbon fiber tube’s curved surface. In the previous design \cite{Wang_IROS-2020}, low-friction plastic bearings snap-fitted into a 3D-printed tube slot constrained the motion. Metal bearings were avoided to prevent wear on the epoxy-based carbon fiber tube, yet epoxy still wears over time, causing pitting and spalling, leading to free play between the bearings and the tube. Additionally, since the foot joint is fixed at the leg’s distal end and swings with the leg, fixing the foot actuator at the base link’s pitch joint creates a rigid, variable-length linkage for torque and velocity transmission, posing a structural design challenge.

These issues highlight the necessity of a fundamentally improved design; accordingly, an extensive literature review was conducted focusing on lightweight, capstan‑driven robots that minimize structural weight by centralizing actuators. A capstan drive uses friction between a rotating drum and a wrapped flexible element (belt, rope, or cable) to transmit force. For example,  the Capler Leg \cite{Hwangbo_IROS-2018} achieves over 96\% energy recuperation during continuous jumping with just 5\% mechanical losses by locating its motor at the hip and using a single capstan drive. Similarly, the LIMS2 (Ambidex) system \cite{Kim_TRO-2017} transfers high gear ratios through capstan cables to distal joints, achieving sub‑human limb inertia and exceptional stiffness.

\begin{figure}
\centering
\includegraphics[width=1.0\textwidth]{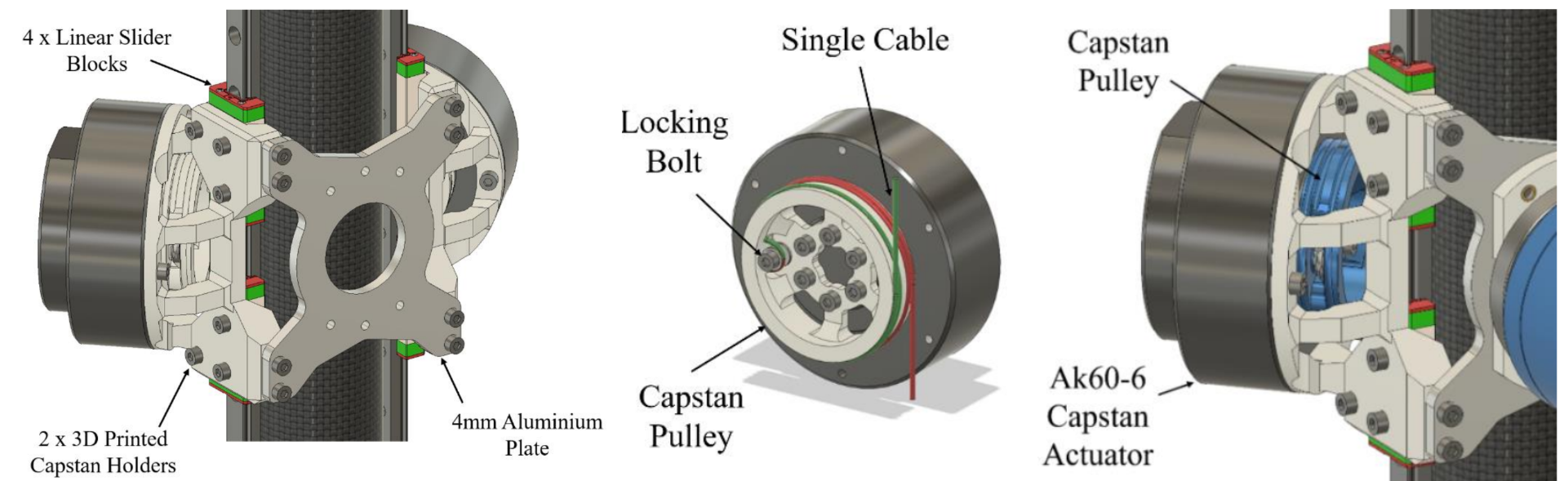}
\caption{Visual illustration of how the differential capstan sliding joint works.}
\label{fig:capstan}
\end{figure}

To adapt SLIDER to the Capstan drive mechanism, a differential drive scheme was proposed, utilizing two identical actuators to control both the foot joint and the sliding joint simultaneously. An illustration of how this differential mechanism works is shown in the Fig.~\ref{fig:differential} below.
\begin{figure}
\centering
\includegraphics[width=1.0\textwidth]{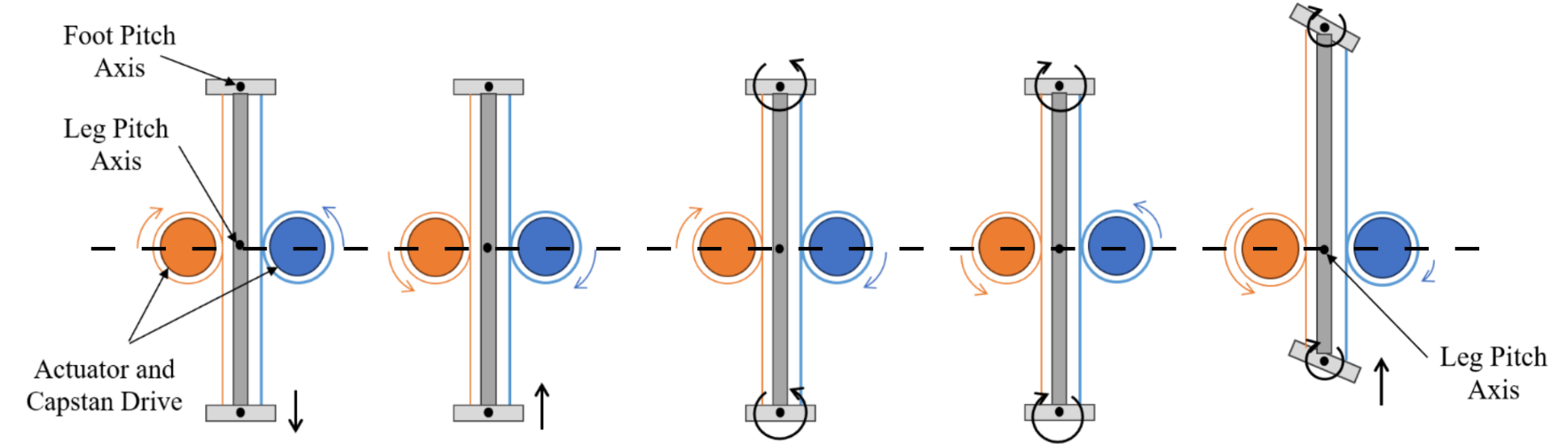}
\caption{Visual illustration of how the differential capstan sliding joint works.}
\label{fig:differential}
\end{figure}
As depicted, each leg assembly features a tightened steel cable on both sides symmetrically, which wraps around a pulley mounted on an actuator, with the endpoints bolted to the feet(shown in Fig.~\ref{fig:capstan}). Although only one foot is necessary, the cable must extend to the top and be anchored to an output lever to ensure proper functionality of the differential capstan drive throughout the entire sliding span. For simplicity, the lower foot design is mirrored at the top. Additionally, the linear motion constrain of the leg is handled by a LML9B metal linear rail, which is tough, smooth, and small enough for this application.
When both leg motors rotate in the same direction at matching speeds, sliding motion is actuated. Conversely, when they rotate in opposite directions, the leg remains stationary while the foot end rotates. The superposition of these two effects, illustrated on the left of the figure, enables simultaneous yet independent control of the prismatic and foot joints. 

With the reduced Degree of Freedom (DoF) in the foot joint and updated controller framework, only linear contact is required. Therefore the new foot design is much more simplified, which can be done using a 12mm aluminum extrusion. The new model weighs only 73.7g, which is just 80.9\% of the previous version's weight. Given the significant improvements outlined above, the new SLIDER design sets a benchmark for future hardware iterations and serves as a potential testing platform for the new control software framework. A complete CAD rendering and an assembled photo are shown in Figure~\ref{fig:combined} in the Introduction.

\section{DRL Control Policy based on eHZD}

\subsection{Extended Hybrid Zeros Dynamics}

HZD has been widely applied for underactuated robot locomotion,  which is especially suitable for point or line contact locomotion. However, the subsequent result is valid only under the condition that all joints are revolute\cite{HZD0}:
\begin{itemize}
    \item \textbf{Existence:} There is a unique limit cycle in an embedded manifold that satisfies the virtual constraints in the state space.
    \item \textbf{Convergence:} A control law can be devised that ensures that all points in the neighborhood of this limit cycle converge asymptotically to the curve.
\end{itemize}

Here, the limit cycle is a closed trajectory in phase portrait. HZD cannot be directly applied to robots with prismatic joints, as the existence of a limit cycle is not guaranteed in such systems. Therefore, we propose the extended HZD method as shown in Fig.~\ref{fig:HZD0.png}:

\begin{figure}[h]
\centering
\includegraphics[width=0.8\textwidth]{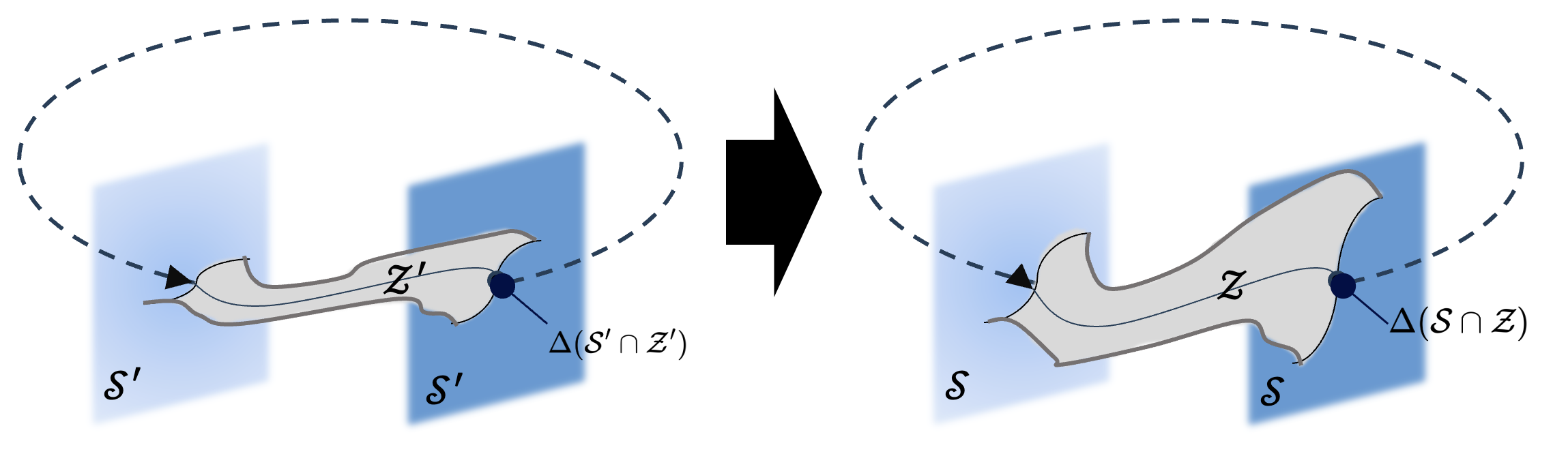}
\caption{Transformation of Zero Dynamic space}
\label{fig:HZD0.png}
\end{figure}

Inside this figure, $S$ is the state space where the impact happened, and $Z$ is the invariant set based on Zero Dynamics assumption. $\Delta(\mathcal{S}\cap\mathcal{Z})$ means the discrete transformation when impact occurs. The left figure expresses the HZD applied in robot model with a virtual knee, and the right figure is the robot model with prismatic joints. The methods of mapping from left to right can be expressed by the following Fig.~\ref{fig:HZD1.png}.
\begin{figure}[h]
\centering
\includegraphics[width=0.8\textwidth]{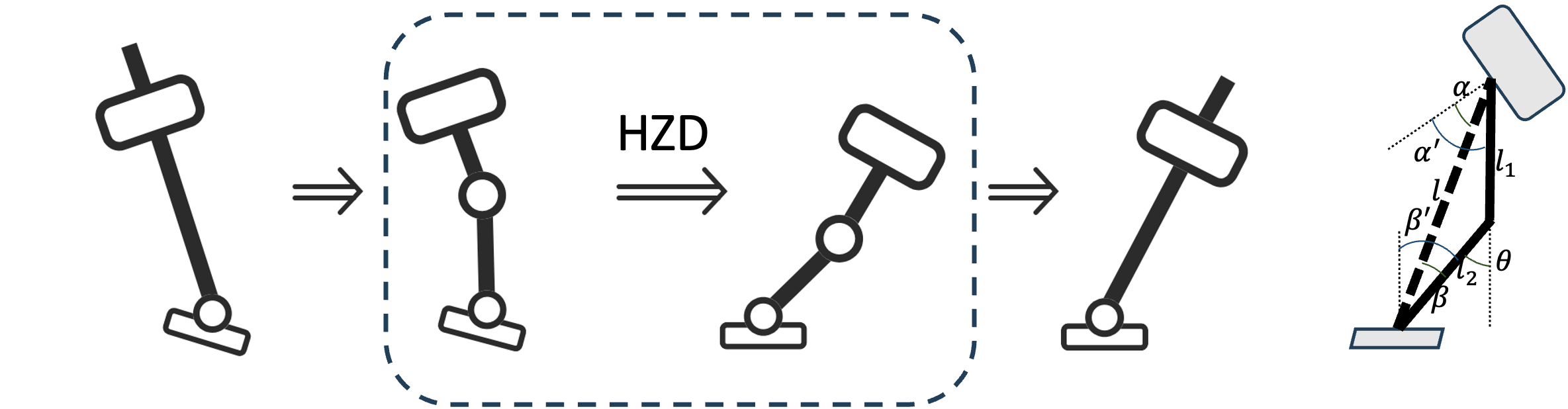}
\caption{Mapping between prismatic model and knee model}
\label{fig:HZD1.png}
\end{figure}

As shown in the Fig.~\ref{fig:HZD1.png}, we first approximate the robot model of the prismatic joint into a virtual model with a revolute knee, varying the distance between the foot and hip to simulate the variation of the linear direction. According to the comparison between the masses of the components in Chapter 2, we assume that the masses of the legs and feet are negligible compared to hip weight. In this way, the phase portrait in the knee model can be converted to the phase portrait for the prismatic model. The corresponding transformation relation between position and velocity is:

\begin{equation}
\begin{bmatrix}
q_{\alpha}\\[3mm]
q_{\beta}\\
\end{bmatrix} \;=\;
\begin{bmatrix}
1 & 0 &-\frac{l_{2}}{l} \\[3mm]
0 & 1 &-\frac{l_{1}}{l}
\end{bmatrix}
\,{\mathbf{q}}'_{T};\    \   \
\\
\dot{\mathbf{q}}_{T} \;=\;
\begin{bmatrix}
1 & 0  & -\frac{l_{2}}{l}\cos(\theta) \\[3mm]
0 & 1  & -\frac{l_{1}}{l}\cos(\theta) \\[3mm]
0 & 0  & -\,2l_{1}l_{2}\cos(\theta)
\end{bmatrix}
\,\dot{\mathbf{q}}'_{T}
\label{eq:1}
\end{equation}
In which the $q_{T}=[q_{\alpha}\ q_{\beta}\ q_{l}]^{T}$ express the position of hip pitch, ankle pitch, and sliding joint. $q'_{T}=[q_{\alpha}\ q_{\beta}\ q_{l}]^{T}$, $q'_{T}=[q'_{\alpha}\ q'_{\beta}\ q'_{\theta}]^{T}$ express the position of hip pitch, ankle pitch and knee joint in virtual robot model. after the prismatic joint position transform $q_{l} = \sqrt{l_1^2+l_2^2+2l_1l_2cos(\theta)}$, equation.~\ref{eq:1} can be applied to transform all state portrait from virtual knee model to the prismatic model.

Thus, once we have obtained the virtual limit cycle with the knee, we can obtain the approximation of the limit cycle in the state portrait of the ideal locomotion gait of the corresponding prismatic joint model. 
To implement HZD as described in \cite{HZD0}, we set up the virtual constraints, adopting the configuration style of\cite{HZD1} to ensure the output matrix is full-rank.

\begin{equation}
L(\theta) 
= \sum_{s \in S} \frac{1}{T}\,
\Bigl(\sum_{t = 1}^{T}\Bigl\|
\tau_{s}\bigl({q}_{t},{\dot{q}}_{t} \bigr)
\Bigr\|^2\Bigr)
\label{eq_2}
\end{equation}
We have used only the equation.~\ref{eq_2} as the cost function to optimize the energy consumption. Here $T$ is the time period for each step, and $\tau$ is the torque output by the control law generated by the HZD method. 



\subsection{Guided DRL structure}

In the simulation, We directly treat the upper foot and the lower foot as independent actuated joints while making the torque output of the upper foot always 0. We use the proximal policy optimization (PPO) algorithm with eHZD gait library for control policy training, and the whole training framework of Guided DRL is shown in Fig.~\ref{fig/DRL1.png}:

\begin{figure}[h]
\centering
\includegraphics[width=0.9\textwidth]{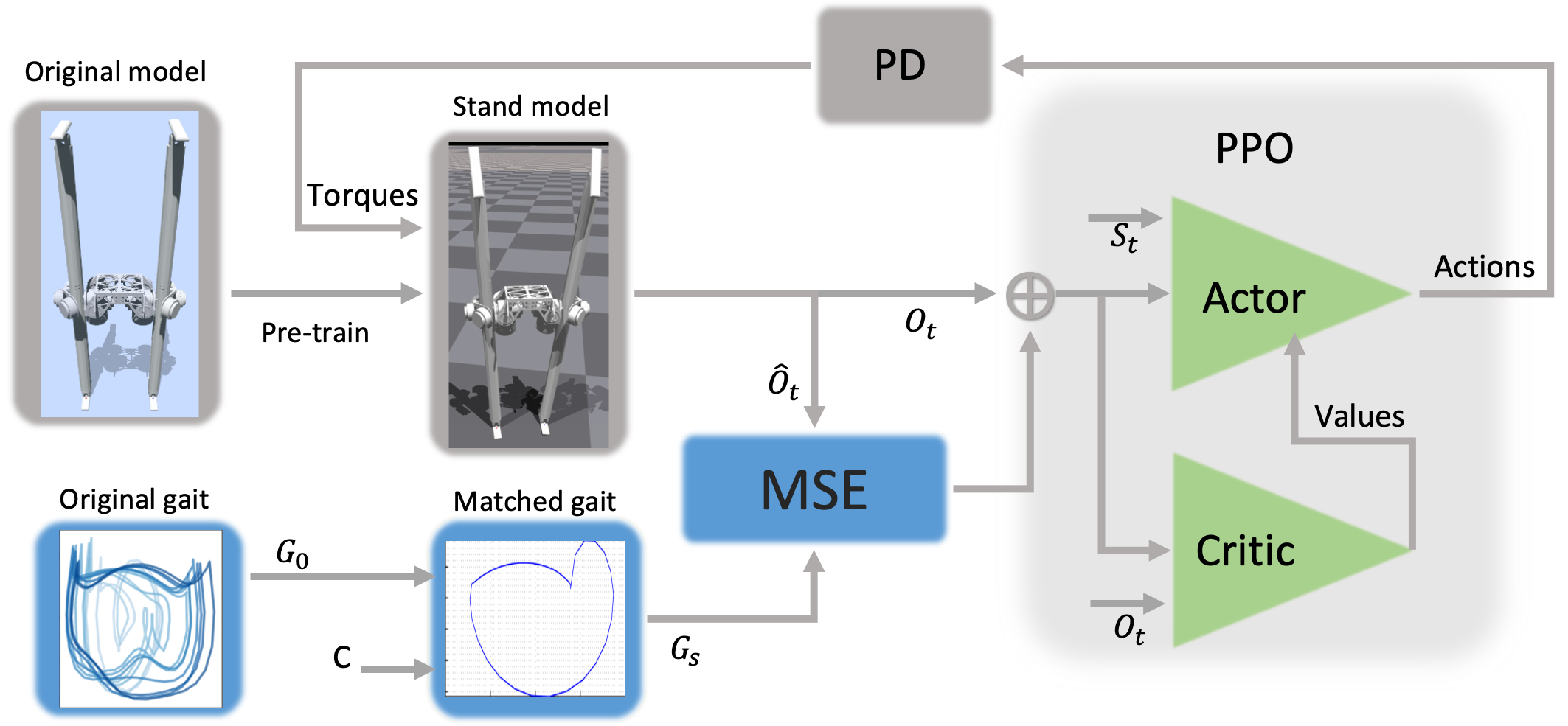}
\caption{The structure of Guided DRL}
\label{fig/DRL1.png}
\end{figure}

Here, the number of environments is $n_{env}$. The observation space is $O_{t}\subseteq\mathbb{R}^{n_{env}\times235}$, which consists of CoM linear velocity space $v_{lin}\subseteq\mathbb{R}^{n_{env}\times6}$, CoM angular velocity space $v_{ang}\subseteq\mathbb{R}^{n_{env}\times6}$, weight projection space $g_{pro}\subseteq\mathbb{R}^{n_{env}\times3}$, CoM velocity command space $C\subseteq\mathbb{R}^{n_{env}\times4}$, actuated state position space $q_{dof}\subseteq\mathbb{R}^{n_{env}\times12}$, actuated state velocity space $\dot{q}_{dof}\subseteq\mathbb{R}^{n_{env}\times12}$, action space $A_{t}\subseteq\mathbb{R}^{n_{env}\times12}$, and a height point cloud $H_{t}\subseteq\mathbb{R}^{n_{env}\times180}$. The actions space is the desired position space for all actuated joints, which will be mapped to the final motor torque space by a PD controller. Here, the Actor network is composed of 3 hidden layers of 256 neurons, which maps the observation space $O_{t}$ to the action space $A_{t}$, and the neural network structure of the Critic network is the same.


We generated a gait library that contains 125 distinct velocity commands. Each gait includes 26 different DoF positions or speeds, and the full-time step length for each periodic walking is 42. In Figure\ref{fig/DRL1.png}, It can be observed that the closest gait space $G_s\subseteq\mathbb{R}^{n_{env}\times42\times26}$ is found for each environment based on its $x$ direction and $y$ direction velocity commands via our original gait library $G_0\subseteq\mathbb{R}^{125\times42\times26}$. At the same time, the original observation space $O_t\subseteq\mathbb{R}^{n_{env}\times235}$ takes the corresponding dimensions to form a new control observation space $\hat{O}_t\subseteq\mathbb{R}^{n_{env}\times26}$, so that the negative Mean Square of error(MSE) can be written as:
\begin{equation}
L_{MSE}(O_{t})
= 
    -E\Bigl(
\bigl\|\hat{O}_t-\hat{G}_t 
\bigr\|\mid \hat{G}_t:=\{ 
g_{i}\in G_{s} \mid argmin(\bigl\|g_{i}-o_{i}\bigr\|)
\}
\Bigr)
\label{eq:2}
\end{equation}

The reward of the whole algorithm can be written as:
\begin{equation}
R_{t}(O_{t}) 
= L_{MSE}(O_{t}) + R_{t}' (O_{t})
\end{equation}

Where $R_{t}$ is the reward function of the whole DRL model, and its composition is shown in the following equation:
\begin{multline}
R_{t}(O_{t}) = L_{MSE}(O_{t}) - 0.5\,\tanh\!\left(\left\|\tau_{t}\right\|\right) -3\tanh\!\left(\left\|v_{ang}\right\|\right) - \tanh\!\left(\left\|v_{lin} - v_{lin\_des}\right\|\right)\\
- \tanh\!\left(\left\|v_{ang} - v_{ang\_des}\right\|\right)+2\tanh\!\left( H_{t} \right)-\tanh\!\left(\left\| A_{t} - A_{t-1} \right\|\right)
\end{multline}

Here $\tau_{t}$ is the torque output of each actuator, $v_{ang}$ and $v_{ang\_des}$ is the actual and desired angular velocity of main body, $v_{lin}$ and $v_{lin\_des}$ are the actual and desired CoM linear velocity, $H_{t}$ is the height of main body and $A_{t}$ is the position command for each joint at time $t$.


Training DRL directly based on the gait library will lead to continuously triggering the falling down termination condition because of small line-feet contact. To reduce such cases, we use the same neural network to pre-train a standing model in advance, remove the reward term of gait, and enlarge the standing rewards. The walking model was initialized with the neural network from the standing model and then trained further following the rest of the DRL framework. The algorithm of the framework is shown in Algorithm 1.

\begin{algorithm}[H]
\caption{}
\SetAlgoLined
\KwIn{Initial policy parameters $\theta$, Gait library $G_{0}$, value function paramerters $\phi$, and other hyperparameters}
\textbf{Pre-train:} $\theta\gets \theta_{standing}$\\
\For{$k=0,1,\dots$}{
    update $C$ \\
    $G_{s}:=\{g_{i}\in G_{0}| argmin(\left \| V(g_{i} )-C\right \|)$\;
    \For{$t=0,1,\dots$}{
         $\hat{G}_t:=\{ 
            g_{i}\in G_{s} \mid argmin(\bigl\|g_{i}-o_{i}\bigr\|)
            \}$\\
         $L_{MSE}(O_{t})
            = 
                E(
            \bigl\|\hat{O}_t-\hat{G}_t 
            \bigr\|)$\\
       
    } 
    PPO: Compute advantage estimates $A_t$ and returns $R_t$ using GAE\\
    \For{$i=0,1,\dots$}{
       $\theta\gets \theta+\alpha \nabla _{\theta}L_{i}^{t}(\theta)$\\
       $\phi\gets \phi-\alpha \nabla _{\phi}L_{i}^{t}(\phi)$       
    }
    
}
\end{algorithm}

\section{Experimental Results and Analysis}

We used an AMD Ryzen 9 CPU and an NVIDIA 4090 GPU for the nonlinear optimisation of eHZD and the training of Guided DRL, respectively. We use Issac Gym\cite{issacgym} for multi-environment parallel training, and the parallel training environment number is 2048. The eHZD optimisation took nearly 20 hours with 125 different speed commands, while the DRL took nearly 10 hours with 40,000 epochs. For eHZD, we used the URDF model with a virtual rotating knee joint, while for the DRL, we used the prismatic model. 

\begin{figure}[h]
\centering
\includegraphics[width=1.0\textwidth]{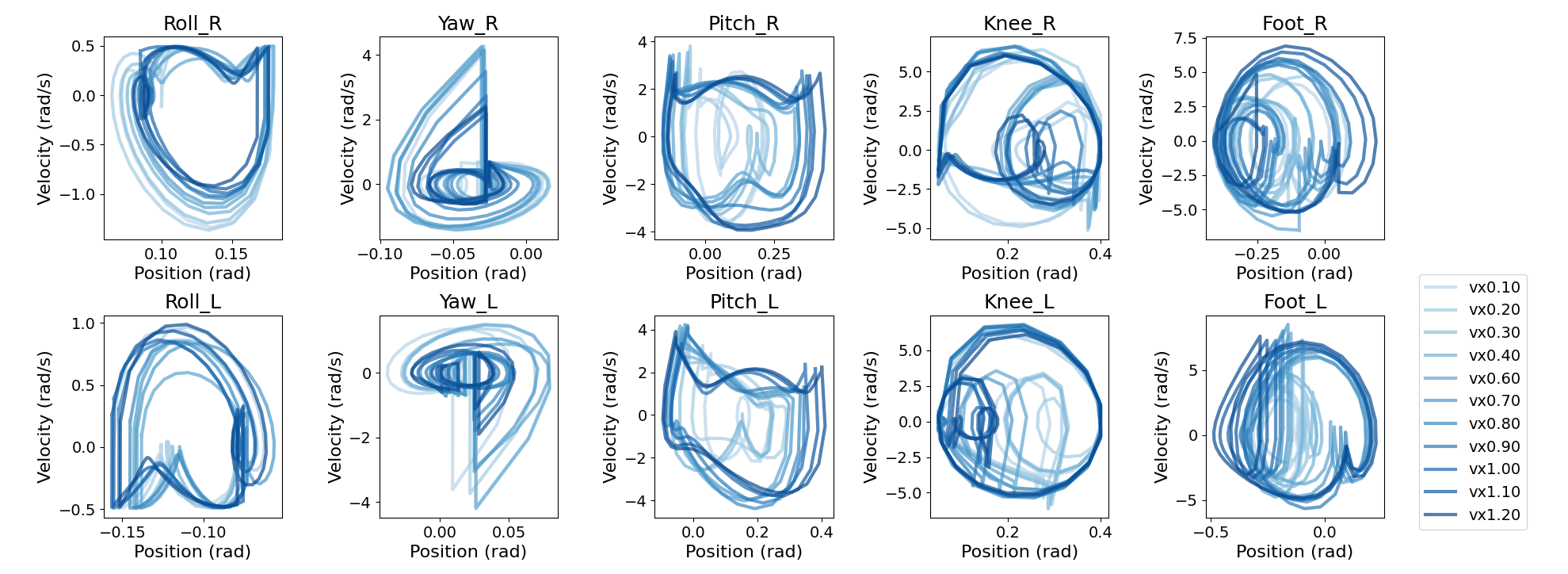}
\caption{Limit Cycles obtained for different forward velocity commands. It indicates that similar command velocities yield similar limit cycles, demonstrating a smooth transition in gait behavior without any abrupt changes.}
\label{fig:HZD2.png}
\end{figure}

We used FROST\cite{FROST} for noise-free ideal nonlinear optimisation at a certain command speed. We performed the ideal limit cycle optimisation within $[-0.1\ m/s , $ $0.1\ m/s]$ in $y$ direction and from $[0.0\ m/s, 1.2\ m/s]$ in $x$ direction at intervals of every $0.05\ m/s$ and obtained a gait library of  velocity and position for all DoF. Some of the state portrait plot in specific commands is shown in Fig.~\ref{fig:HZD2.png}. These results confirm that within the specified velocity ranges, the mapping from the virtual knee's joint space to the prismatic space is smooth and continuous.

In order to verify the performance ability of this locomotion control policy under various speed commands, we set speed tracking tests between $0.0 \ m/s$ to $3.0\  m/s$ and compared it with the combination of Model Predictive Control (MPC) and Whole Body Controller (WBC)\cite{MPC0}. We define "successful balance" as the ability to track the desired velocity continuously for 10 seconds without triggering any termination conditions. It was found that the balancing success rate at each speed, and the actual speed MSE at each speed are shown in the Table.~\ref{table_result}.
\vspace{-1.5em} 
\begin{table}[!ht]
    \centering
    \resizebox{1.0\textwidth}{!}{%
    \begin{tabular}{c|c|c|c|c|c|c|c}
    Command Velocity & 0.2 & 0.4 & 0.6 & 1.0 & 1.2 & 2.2 & 3.0 \\ \hline
    Success rate(Ours):    & $100\%$  & $100\%$ & $100\%$ & $100\%$ & $100\%$ & $100\%$ & $73.1\%$\\
    Success rate(MPC):    & $100\%$  & $90\%$ & $30\%$ & $0\%$ & $0\%$ & $0\%$ & $0\%$\\
    MSE(Ours):& $0.0317$  & $0.0222$ & $0.0229$ & $0.0221$ & $0.0232$ & $0.520$ & $1.6$\\
    MSE(MPC):& $0.00538$  & $0.0251$ & $0.0593$ & Inf & Inf & Inf & Inf\\
    \end{tabular}
    }
\caption{Comparison between our method and MPC}
\label{table_result}
\end{table}
\vspace{-1.5em}

\begin{figure}[h]
\centering
\includegraphics[width=0.85\textwidth]{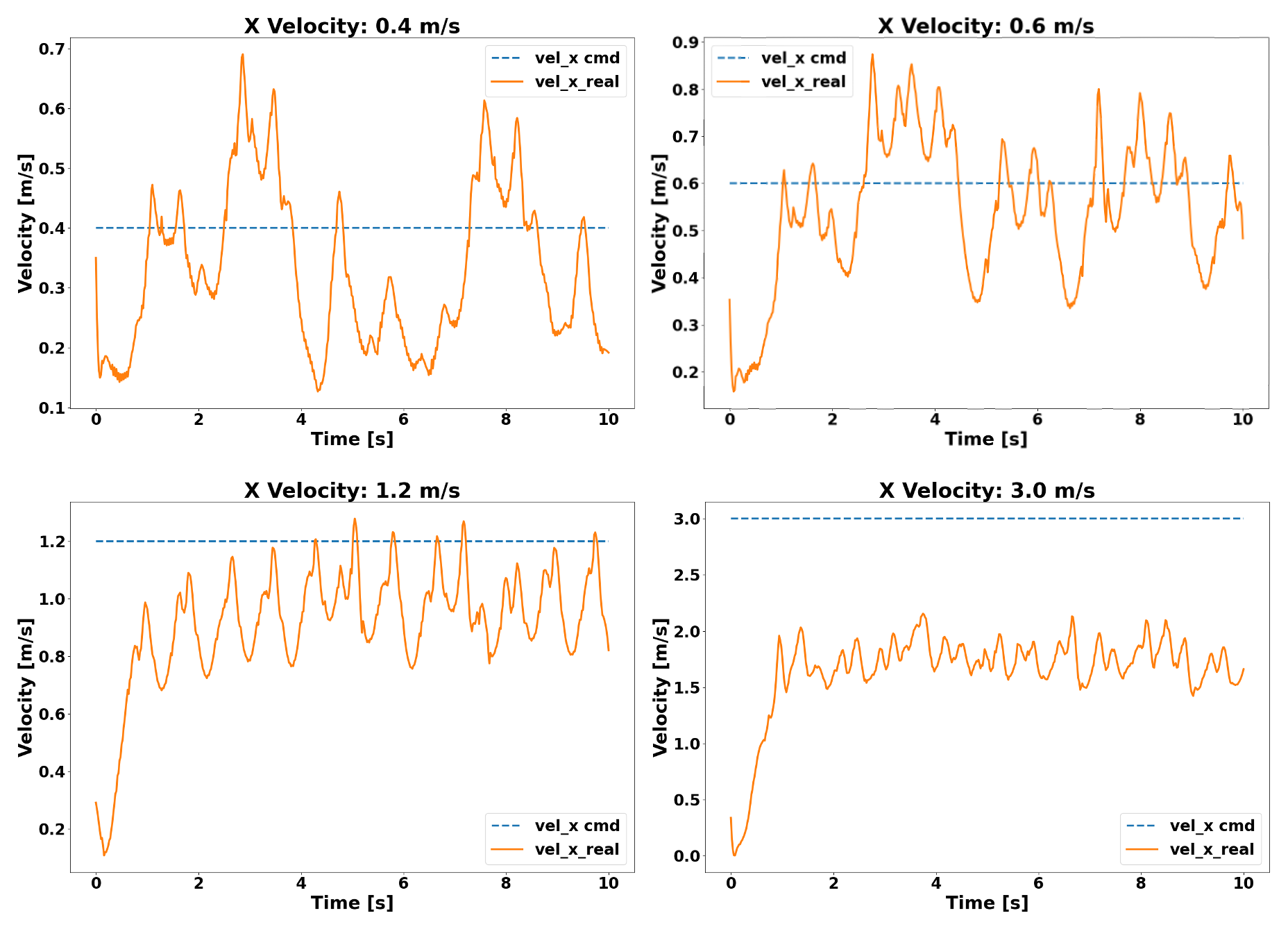}
\caption{The velocity tracking property of four different commands. At $0.4 m/s$, $0.6 m/s$, and $1.2 m/s$, the velocity tracking performance are all qualified, while at $3.0 m/s$ it can only reach $1.7 m/s$ velocity }
\label{fig:result1}
\end{figure}

Ten trials are performed to obtain the experiment results. In each trial, 100 robots are running in parallel with the same controller in the simulation. From here, we can see that although MPC is a little better at slower speeds like $0.2\ m/s$, it will not be able to maintain balance at increased speeds. Also, the performance of both methods at $0.4\ m/s$, $0.6\ m/s$ and $1.2\ m/s$ $3.0\ m/s$ is shown in Fig.~\ref{fig:result1}.

The results indicate that our model has robust performance across a broad range of command velocities, given that the consistent MSE is observed for moderate and high speeds in Table.~\ref{table_result}. The MSE remains nearly constant as the command velocity increases, which means the control policy is well-matched with the gait library for velocities. At low command velocities, such as $0.4m/s$, we observe a slightly worse tracking performance compared to MPC; this issue presents an opportunity for further refinement. Moreover, when the command velocity reaches $3.0 m/s$, the robot is unable to reach the desired speed, indicating an upper limit in the current gait formulation for high-speed locomotion.


\section{Conclusion}
In this paper, we enhanced the flexibility of the previous SLIDER robot by replacing the planar contact constraints with line contact, thereby increasing the upper-speed limit while preserving the prismatic joint configuration. At the same time, the addition of the Yaw joints and the optimized mass distribution design facilitates the construction of the control policy in the later stage. We proposed a gait library for eHZD and designed a Guided DRL framework for our SLIDER to ensure its reliability. The motion limit speed is increased by 150\% compared to the MPC approach on the previous version of SLIDER. 

In the future, we plan to combine this framework with the MPC framework to ensure that this robotic system can reduce the gap of simulation to the real world when transferring to real life. 

%
%

\end{document}